# Ensemble BERT: A student social network text sentiment classification model based on ensemble learning and BERT architecture.

**Kai Jiang[1], Honghao Yang[2], Yuexian Wang[3], Qianru Chen[4], Yiming Luo[5*]**

[1]Business college, Southwest University, Chongqing, 400715, China
[2]Foreign languages college, Central China Normal University ,Wuhan, 430079 China
[3]School of information and business management, Dalian Neusoft University of Information，Dalian Liaoning, 116000, China
[4]School of Art, Design and Fashion, Zhejiang University of Science and Technology, Hangzhou, 310000, China
[5]Business school University of Sydney, Sydney, New South Wales, 2150, Australia

[*]Yiming Luo is the corresponding author and responsible for the design of the article. The corresponding author email is yluo5754@uni.sydney.edu.au

**Abstract.** The mental health assessment of middle school students has always been one of the focuses in the field of education. This paper introduces a new ensemble learning network based on BERT, employing the concept of enhancing model performance by integrating multiple classifiers. We trained a range of BERT-based learners, which combined using the majority voting method. We collect social network text data of middle school students through China's Weibo and apply the method to the task of classifying emotional tendencies in middle school students' social network texts. Experimental results suggest that the ensemble learning network has a better performance than the base model and the performance of the ensemble learning model, consisting of three single-layer BERT models, is barely the same as a three-layer BERT model but requires 11.58% more training time. Therefore, in terms of balancing prediction effect and efficiency, the deeper BERT network should be preferred for training. However, for interpretability, network ensembles can provide acceptable solutions.

## 1. Introduction

Research based on neural network technology has gradually become more and more widely used in many fields [1]. With the rapid growth of Deep Learning, especially in the field of Natural Language Processing (NLP), a growing number of researchers are exploring the implementation of deep networks for NLP tasks, especially text classification [2]. Deep networks, such as the transformational BERT, have achieved unprecedented performance in several NLP tasks. However, significantly increasing the depth of networks often results in a substantial rise in computational costs, which is particularly problematic in resource-limited scenarios. Additionally, deep networks face certain barriers in terms of interpretability [3]. Therefore, this study introduces a novel ensemble learning framework that combines the powerful semantic capturing ability of transformers with the efficiency of ensemble learning. Ensemble learning can increase robustness and reduce bias, while shallow networks can facilitate interpretive analysis.

This paper aims to investigates the following questions, and the N is the number of single-layer BERT base model:

**Question 1**: Can an ensemble of N single-layer base models achieve good results in student sentiment classification?

**Question 2**: Does the predictive ability of an ensemble model with N single-layer base models surpass that of training a single N-layered BERT model?

**Question 3**: Is the training time for an ensemble of N single-layer base models shorter than training a single N-layered BERT model when there is equal predictive accuracy?

## 2. Dataset

We demonstrate the effectiveness of our approach by applying it to sentiment trend analysis tasks on social network text. The single-layer decoder architecture of the BERT model as the base model of the ensemble is used, and the classification results are output using an average voting method. To simplify the experiment, we select N(number) as 3. The data is social network data from students in three middle schools in Xiangtan City, Hunan Province, China. Weibo is the largest social network platform in China. After obtaining students' consent, students uploaded their Weibo accounts anonymously. We collected the account data of 324 students and captured 100 Weibo text contents of each student in the last 3 years. If there are not 100 pieces of content, all tweet content will be used. Finally, a total of 30012 pieces of data were included. The students and their parents approved and consented to this data, and the students and psychological education experts double-checked the data annotation. Next, we will present related work, followed by an explanation of the model architecture. Then, we will present and analyze the experimental results.

## 3. Related work

Traditional text vectorization methods such as one-hot encoding, bag-of-words, and TF-IDF models have inherent limitations [4]. They often lead to sparse text representations, resulting in significant computational overhead and insufficient understanding of contextual semantic relationships. These limitations prevent them from solving the challenges posed by polysemy [5]. However, extracting text features through neural network models can effectively resolve these semantic issues in text features, effectively addressing the challenges of polysemy and language Ambiguity in text classification [6]. Particularly, the use of pre-trained models enhances the acquisition of text semantic representations. Widely used neural network-based text representation models are modelled based on the relationship between context and target words. The commonly used models include the pre-training model based on the static word embedding method like Word2Vec and the BERT model, which based on the dynamic word embedding method.Word2Vec is a neural network that converts each word segment (token) in text data into a vector in k-dimensional space by training the network on a given corpus[7].Training models based on Word2Vec include the Skip-gram model, which can predict the context of a known target word or token[6].BERT is a pre-training model based on deep learning for language representation proposed by Google in 2018[8]. It is also one of the most important pre-training models in NLP now. It uses a bidirectional encoder structure, so it can well capture the differences between text contexts. semantic relationship, and our base model comes from this. The structure of BERT is shown in Figure 1. The first layer is the word embedding layer. After word embedding, it enters 12 transformer layers and obtains the final feature representation.

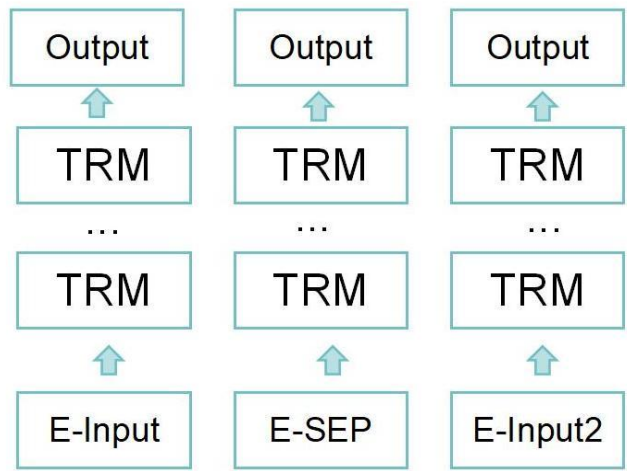

**Figure 1.** The structure of BERT.

The embedding layer structure of the BERT model, shown in Figure 2, consists of word, segment, and position vectors that add together to create new comprehensive embeddings. Word embeddings map each word or sub-world unit to a vector in a large-dimensional space. Segment embeddings enable BERT to distinguish and process individual texts or pairs of texts. Positional embeddings are essential for the Transformer architecture utilized in BERT, as they provide sequential information and enable the model to understand the position of words in a sentence. The embeddings are non-recursive, which makes them crucial for the architecture.

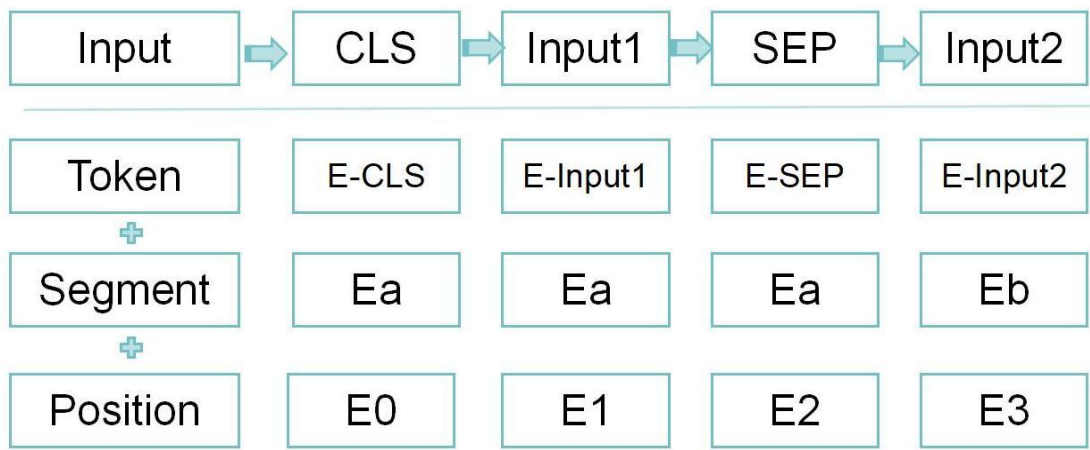

**Figure 2.** The structure of BERT embedding layer

After entering the first layer of analysis, data that was previously labeled incorrectly is utilized as a fresh training set for preliminary instruction. In this training phase, 15% of the words in the provided word sequence are randomly masked, and then predictions are made for the masked words. The strategy for masking includes substituting the chosen word with the mark 'MASK' token 80% of the cases, with a random word 10% of the cases, and leaving the chosen word unchanged 10% of the cases, to resemble the actual observed words more closely. Once the pre-training tasks are completed, the BERT model's representation of the input sentences can be obtained. This involves using the output of the last layer of the BERT model, typically the researchers employ the output corresponding to the

[CLS] token, as the feature representation because we believe it has included the major information of sentences. These features are then fed into a multilayer perceptron (MLP) for the final output classification in terms of the practical request.

Ensemble learning is a widely used technique in machine learning that improves the performance or robustness of a machine learning model by combining predictions from multiple identical or different models. This includes methods such as maximum voting, averaging, stacking, mixing, bagging, and boosting [9].

## 4. Model architecture and experimental process

We use a single-layer BERT network as the base model, integrate three base models, and finally output the final classification based on the majority voting principle. The structure of the model is shown in figure 3.

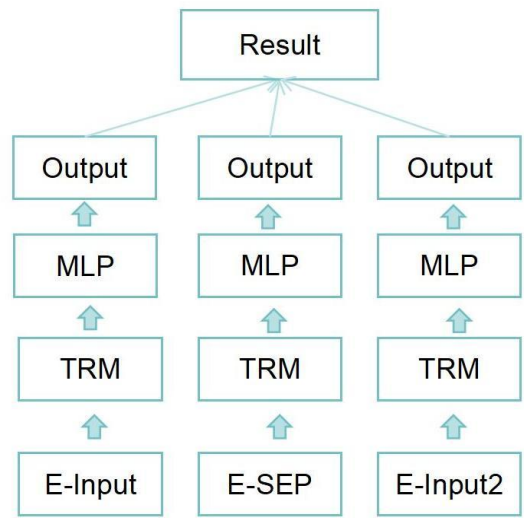

**Figure 3.** The structure of ensemble BERT

The specific steps of the experiment are as follows: Python is used for the experimental processing, and pandas are used to read the data, which contains two columns: one for the review texts and the other for the sentiment labels. The review texts are then tokenized using the Chinese BERT tokenizer (Bert-base-Chinese). The tokenized inputs and sentiment labels are then converted into tensors and a tensor dataset is created. This dataset is then divided into training and validation sets. Three basic BERT models, each with a single hidden layer, are looped and trained. Each model is independently trained using the train dataset and evaluated using the validate dataset. The weights of the models are initialized with the built-in initial weights of the BERT model, and the order of the data batches is randomized in each epoch during training. The predictions of all models are then combined using the majority voting method to determine the final prediction labels. Metrics such as the accuracy, precision, recall, F1 score, and confusion matrix of the ensemble prediction are calculated. For comparative research, a three-layer BERT network and a full 12-layer BERT are also trained for text classification, using an MLP for the output layer. The training times for these models are recorded using the tqdm [10] toolkit. The training is conducted on a CPU, specifically a Ryzen 5800H, and each model is fixed to train for three epochs.

## 5. Model results

To answer the **question 1**, we compare the prediction effect of the base model and the prediction effect of ensemble learning. The prediction effect of ensemble learning is shown in Table 1, and the prediction effect of the base model is shown in Table 2.

Table 1. Model evaluation for ensemble BERT

| Evaluation index | Value |
|---|---|
| Accuracy | 0.9702 |
| Precision | 0.9870 |
| Recall | 0.9529 |
| F1-score | 0.9697 |

Table 2. Model evaluation for base model BERT

| Evaluation index | Value |
|---|---|
| Accuracy | 0.9612 |
| Precision | 0.9825 |
| Recall | 0.9510 |
| F1-score | 0.9665 |

From the results, the base model achieved high accuracy and other indicators. This is due to the powerful decoder architecture of the BERT model. The ensemble model has improved the accuracy of the base model to more than 97%. When comparing various evaluation indicators between the integrated BERT model and the basic BERT model, the largest percentage gap occurs in the accuracy rate, with a gap of approximately 0.94%. This is already a relatively significant gap in a fairly accurate model. Therefore, we believe that the integrated model does improve experimental results.

To answer the **question 2 and 3**, the result for BERT's model with three layers is shown in Table 3.

Table 3. Model evaluation for 3-layers BERT

| Evaluation index | Value |
|---|---|
| Accuracy | 0.9707 |
| Precision | 0.9937 |
| Recall | 0.9470 |
| F1-score | 0.9699 |

When comparing various evaluation indicators between the 3-layer BERT model and the integrated BERT model, the largest percentage gap occurs in precision. The difference is about 0.68%, and the accuracy difference is only 0.05%, so we assume the prediction performance of these two methods is basically the same. However, it can be seen from Table 4 that under the same effect, the training time of ensemble learning exceeds the training time of deep network by about 11.58%. Therefore, under the same circumstances, choosing a deeper neural network may be a better choice.

Table 4. Model training time results

| Model | Training Time (min) | Accuracy | Accuracy per min |
|---|---|---|---|
| Ensemble BERT | 212 | 0.9702 | 0.0046 |
| 3-layer BERT | 190 | 0.9707 | 0.0051 |
| Standard BERT（12 layers） | 792 | 0.9982 | 0.0013 |

When comparing our final analysis of the confusion matrix for the ensemble model, the results are as follows: True Negative is 11858, False Positive is 150, False Negative is 565, True Positive is 11427. We observed that the model had significantly more False Negatives than False Positives, indicating that it more often misclassified posts that were positive in sentiment as negative. Additionally, the content of students' social media may include Internet slang and culturally specific ways of expressing emotions, which may challenge the model's ability to accurately identify sentiment, resulting in a high rate of False Negatives.

## 6. Conclusion

This paper presents a novel ensemble learning network based on BERT. We trained a set of BERT-based learners and combined them using the majority voting method. We proposed three questions and applied this method to classify emotional tendencies in the social network texts of middle school students. Experimental results indicate that ensemble learning shows some improvement over the base model, but the performance of the ensemble learning model composed of three single-layer BERT models is roughly equivalent to that of a three-layer BERT model, although it requires additional training time. Therefore, in terms of predictive effectiveness, deeper networks should be considered the preferred choice. However, in terms of interpretability, an ensemble of shallow networks may provide an acceptable solution. In addition, a limitation of this model is that it only discusses the case where N equal to 3. Future research could explore the optimal balance between time and efficiency for different numbers of layers and discuss the impact of different base models on sentiment classification of student texts.